\pdfoutput=1

\documentclass[a4paper]{article}

\usepackage{INTERSPEECH2022}
\usepackage{latexsym}
\usepackage[utf8]{inputenc}
\usepackage{microtype}
\usepackage{graphicx}
\usepackage{booktabs}
\usepackage{multirow}

\usepackage{rotating}
\usepackage{xcolor}
\usepackage{url}
\usepackage{comment}
\usepackage{placeins} 

\title{ASR-Generated Text for Language Model Pre-training \\
Applied to Speech Tasks}

\name{Valentin Pelloin$^1$, Franck Dary$^2$, Nicolas Hervé$^3$, Benoit Favre$^2$, \\ Nathalie Camelin$^1$, Antoine Laurent$^1$, Laurent Besacier$^4$}
\address{
  $^1$Laboratoire d’Informatique de l’Université du Mans (LIUM), France\\
  $^2$Aix Marseille Univ, Université de Toulon, CNRS, LIS, Marseille, France \\
  $^3$Institut National de l'Audiovisuel (INA), France \\
  $^4$Naver Labs Europe (NLE), France}

\email{valentin.pelloin@univ-lemans.fr}

\begin{document}

\maketitle

\begin{abstract}
We aim at improving spoken language modeling (LM) using very large amount of automatically transcribed speech. We leverage the INA (French National Audiovisual Institute\footnote{\url{https://www.ina.fr}}) collection and obtain 19GB of text after applying ASR on 350,000 hours of diverse TV shows. From this, spoken language models are trained either by fine-tuning an existing LM (FlauBERT\footnote{\url{https://github.com/getalp/Flaubert}}) or through training a LM from scratch.
New models (FlauBERT-Oral) are shared with the community and evaluated for 3 downstream tasks: spoken language understanding, classification of TV shows and speech syntactic parsing. Results show that FlauBERT-Oral can be beneficial compared to its initial FlauBERT version demonstrating that, despite its inherent noisy nature, ASR-generated text can be used to build spoken language models.


\end{abstract}

\section{Introduction}

Large pretrained language models (LMs) have been widely adopted as backbone for performing natural language processing tasks, whether on text~\cite{koroteev2021bert} or on speech transcripts~\cite{qin2021survey}. Those models are typically trained with massive quantities of text, and do not reflect well the specifics of spoken language. In this study, we investigate whether there is some benefit in better covering spoken-language phenomena in pretrained LMs when applying them to speech use cases.

Pretrained LMs, such as BERT~\cite{devlin2018bert}, have been applied with success to a range of tasks that involve speech: ASR hypothesis rescoring~\cite{shin2019effective}, Spoken Language Understanding~\cite{liu2020jointly}, Dialog State Tracking~\cite{lai2020simple}, Conversation Summarization~\cite{ma2021extractive}, Emotion Recognition~\cite{siriwardhana2020jointly}, Speaker and Language recognition~\cite{ling2020bertphone}. Yet, LM training data is dominated by textual web content, which diverges from actual spoken language. Even though the web is filled with conversation-like interactions, such as forums and social media in which agrammaticality or colloquiallity are frequent, it does not include some fundamental properties that are inherent to spontaneous speech: floor holding and grabbing, conversational discourse markers, disfluencies such as repairs, adverse effects of noise and speaker overlap, among others. One may wonder if addressing such phenomena without having seen them in training is a capability of current techniques associated with LMs, and whether they matter in the context of everyday spoken language processing tasks.

Building LMs aware of speech phenomena is a difficult task. Recent work addresses the problem through direct learning of acoustic representations~\cite{baevski2020wav2vec,hsu2021hubert}, but have yet to show competitive performance compared to word-based models for high-level linguistic tasks. Instead, we aim at retraining LMs from speech using word transcripts, and are faced with the inherent challenge of gathering enough training material to be on par with text-only models.
Therefore, we investigate the use of massive amounts of automatic speech recognition (ASR) generated text for spoken language modeling. This could bring diversity (oral/spontaneous style, different topics) to the LM training data and  might also be useful for languages with fewer text resources but  high availability of speech recordings. 
More precisely, we build and share oral LMs trained from a large amount (350,000 hours) of French TV shows, and evaluate them on three spoken language tasks: spoken language understanding,  topic classification of TV shows, and spoken language syntactic parsing. Results suggest that ASR-generated text is a viable alternative to text not originating from speech when training spoken language models.

To our knowledge, this is the first time a LM is trained on such a massive amount of ASR transcripts. Indeed, \cite{kumar-etal-2021-bert} evaluated BERT based language models (BERT, RoBERTa) trained on spoken transcripts to investigate their ability to encode properties of spoken language, but most of their experiments are on small quantities of manual transcripts, except one which is trained on 2000h of ASR output (1k Librispeech + 1k proprietary dataset).
We train LMs on larger quantities of ASR output, analyse their benefits on speech-specific tasks, and make them available for further research.


\section{From FlauBERT to FlauBERT-Oral}

\subsection{Large speech corpus with ASR transcripts}

The FlauBERT-Oral model is trained on 350,000 hours of speech collected from various French TV and radio channels collected at Institut National Audiovisuel (INA) from 2013 to 2020.
TV news channels were continuously captured between 6am and midnight each day (BFMTV, LCI, CNews, France 24, France Info).
For radio, the morning news were used (Europe1, RMC, RTL, France Inter) and for generalist TV channels we collected evening news (TF1, France 2, France 3, M6). 

Collected speech is automatically transcribed with an ASR system built from Kaldi~\cite{povey2011kaldi}. 
In that system, acoustic models are trained from French TV and RADIO corpora (ESTER 1\&2 \cite{galliano2009ester}, REPERE \cite{giraudel2012repere}  VERA \cite{goryainova2014morpho}).
A regular backoff n-gram model is estimated using manual transcripts augmented with several French newspapers \cite{deleglise2009improvements}, resulting in a vocabulary of 160k most frequent words.
Automatic speech diarization of the collection was performed using LIUMSpkDiarization~\cite{meignier2010lium}.
The whole pipeline obtains the following ASR performance on different reference test corpora: REPERE (12.1\%WER), ESTER1 (8.8\%WER) and ESTER2 (10.7\%WER).

As the ASR system outputs raw text without punctuation nor capitalization, speech turns  from diarization serve as pseudo sentences for later processing.
This results in 51M unique speech segments for a total of 3.5G words (19GB of data). It is important to note that this 'oral' data is highly biased towards radio and TV news type.





\subsection{Fine-tuning or re-training FlauBERT}
\label{sec:flaubert-finetuning}
All the models generated in this study are built on the FlauBERT family of models, one of the first pre-trained LMs for French~\cite{flaubert}.
FlauBERT is trained with a masked language modeling (MLM) objective, and implements the RoBERTa architecture \cite{DBLP:journals/corr/abs-1907-11692}. It is trained on 71GB of natural texts collected from various sources, among which a majority of crawled web pages. The baseline for our experiments is \textit{flaubert-base-uncased} (\textbf{FBU})\footnote{\url{https://huggingface.co/flaubert/flaubert_base_uncased}} since we only have lowercase transcripts. 
We compare it to several learning configurations based on the same architecture in order to
 observe the impact of different parameters on the quality of the obtained models. 
 
The first configuration, \textit{FlauBERT-O-base\_uncased} (\textbf{FT}),  consists in
fine-tuning \textit{flaubert-base-uncased} for several epochs using the large corpus of ASR transcripts.

The second configuration \textit{FlauBERT-O-mixed} (\textbf{MIX}) is a full model re-trained using a balanced mix of ASR transcripts and written text. Written texts come from two main sources: a French Wikipedia dump and press articles captured by the OTMedia research platform \cite{otmedia} (online press and AFP agency for the same time period). Overall, this learning dataset is also strongly news-biased.
The written part of this training dataset is 94M randomly selected sentences representing 13G of data stripped from punctuation and capitalization to make it consistent with ASR data. For this mixed model, we also retrain the BPE tokenizer (50K sub-word units).

The third configuration, \textit{FlauBERT-O-asr}, consists in re-training two LMs from scratch using ASR data only. 
For the first model (\textbf{ORAL}), we use the tokenizer provided with \textit{flaubert-base-uncased} and for the second one (\textbf{ORAL\_NB}) we retrain a BPE tokenizer (50K sub-word units).
Both tokenizers share 52\% of their tokens only.

The four models were trained on a single server with two 12-core Xeon CPUs, 256 GB of RAM and 8 Nvidia GeForce RTX 2080 Ti GPUs with 11 GB of memory. It lasted 15 days for 50 epochs of each model in the flaubert-base configuration (137M parameters) using the original FlauBERT code. The models are made available online.\footnote{\url{https://huggingface.co/nherve}}
The following sections describe experiments on various downstream tasks.



\section{Spoken Language Understanding}
Spoken Language Understanding (SLU) systems extract semantic information from  speech utterances \cite{TurDeMori2011}, which can be represented with slots or concepts tags on the token sequence.
Pipeline SLU systems typically generate transcripts from spoken dialogues with an ASR system, and recognize slots with a Natural Language Understanding (NLU) module that only uses the word sequence~\cite{6424196,6998838}.
For example, in the recent approach by~\cite{Ghannay2021}, a first ASR model finetuned on the domain obtains transcriptions, which is fed to a finetuned BERT-like model for classification.
In order to reduce the impact of error propagation in the pipeline, alternative approaches have been proposed, such as end-to-end models that directly predict concepts from speech \cite{DBLP:conf/icassp/DinarelliKJB20,Pelloin2021,Serdyuk2018}.
Our assumption with FlauBERT-Oral models is that the use of ASR-robust pretrained language models for NLU can decrease the impact of ASR errors without resorting to end-to-end training.

\subsection{The MEDIA SLU benchmark}


The French MEDIA dataset \cite{BonneauMaynard2005} contains $1,250$ telephone dialogues captured in Wizard-of-Oz settings for hotel booking and information requests in French. 
Transcribed segments are annotated with $76$ concept tags that denote both domain-specific information (such as \textit{hotel-services} or \textit{room-type}) and generic concepts (like logical connectors as \textit{connectprop}).
Normalized concept values also have to be predicted.
An example can be found in Table \ref{tab:sample-MEDIA}.

\definecolor{color_concept_a}{RGB}{2, 62, 138}    
\definecolor{color_concept_b}{RGB}{230, 57, 70}   
\definecolor{color_concept_c}{RGB}{130, 173, 103} 
\definecolor{color_concept_d}{RGB}{0, 127, 127}   

\newcommand{\openconcept}[2]{\textcolor{#1}{
    \textbf{\small{\textless}}\hspace{-2pt}
    \textbf{#2}
}}
\newcommand{\closeconcept}[1]{\textcolor{#1}{
    ~\textbf{\small{\textgreater}}\hspace{-2pt}
}}

\newcommand{\concept}[3]{\openconcept{#1}{#2}#3\closeconcept{#1}}

\begin{table}[]
\centering
\caption{Sample instance from MEDIA. (a) corresponds to the transcribed sentence; (b) with associated concepts; (c) with concepts and normalized values; (d) concepts in BIO format; utterances were translated to English.}
\begin{tabular}{l p{0.85\linewidth}}
(a) & i would like to book one double room in paris \\ \midrule
(b) & \concept{color_concept_a}{cmd-task}{i would like to book}~~\concept{color_concept_b}{nb-room}{one}~~\concept{color_concept_c}{room-type}{double room}~~in~~\concept{color_concept_d}{loc-city}{paris} \\ \midrule
(c) & \concept{color_concept_a}{cmd-task}{booking}~~\concept{color_concept_b}{nb-room}{1}~~\concept{color_concept_c}{room-type}{double}~~in~~\concept{color_concept_d}{loc-city}{paris} \\ \midrule
(d) & \textbf{\textcolor{color_concept_a}{B-cmd-task I-cmd-task I-cmd-task I-cmd-task I-cmd-task} \textcolor{color_concept_b}{B-nb-room} \textcolor{color_concept_c}{B-room-type I-room-type} O \textcolor{color_concept_d}{B-loc-city}}
\end{tabular}
\label{tab:sample-MEDIA}
\end{table}

We use the official train/dev/test split of the MEDIA corpus, containing respectively $12.9$k, $1.3$k and $3.5$k user utterances.
SLU performance is evaluated with Concept Error Rate (CER), the rate of errors between force-aligned reference and hypothesis concepts, and  Concept-Value Error Rate (CVER), the same for (concept, value) pairs.
Values are extracted with the same rule-based system as in other studies on MEDIA \cite{Ghannay2021,Pelloin2021,hahn2010comparing}.

\subsection{Experiments}
We use the different pre-trained models presented in section \ref{sec:flaubert-finetuning} as pre-trained Natural Language Understanding (NLU) components.
We add a decision layer at the representations generated by the models, in order to tag each input token to its corresponding BIO concept.
Each model is trained for $300$ epochs, and the best checkpoint is selected according to CER on the development corpus.
We release models and make them available online.\footnote{\url{https://huggingface.co/vpelloin/MEDIA_NLU_flaubert_uncased}, \url{https://huggingface.co/vpelloin/MEDIA_NLU_flaubert_finetuned}}

Table \ref{tab:slu-results} presents the \textit{test} results for the different FlauBERT configurations.
The baseline State-Of-The-Art (SOTA) results \cite{Ghannay2021} are obtained with a cascade composed of wav2vec 2.0 \cite{NEURIPS2020_92d1e1eb} (finetuned on CommonVoice \cite{ardila-etal-2020-common} and MEDIA), followed by a finetuned CamemBERT \cite{martin-etal-2020-camembert}, similar to our approach.
For transcription, we use a Kaldi system that reaches $9.1\%$ WER on MEDIA (test)---ASR for the SOTA system has a WER of $8.5\%$.

\begin{table}[]
\centering
\caption{SLU Task: MEDIA test results according to Concept Error Rate and Concept-Value Error Rate (lower is better), using ASR transcriptions as input of NLU FlauBERT models. Confidence intervals are computed using the Student's t-distribution at $95\%$.}
\begin{tabular}{l@{ }cc}
\toprule
\textbf{Model}      & \textbf{\%CER} & \textbf{\%CVER} \\ \midrule
w2v 2.0+CamemBERT \cite{Ghannay2021}       &  $11.2$ &  $17.2$ \\ \midrule
FlauBERT cased   & $13.20 \pm 0.71$ & $18.25 \pm 0.81$ \\
FlauBERT uncased (FBU) & $12.40 \pm 0.69$ & $17.49 \pm 0.79$ \\ \midrule
 FT                  & $\textbf{11.98} \pm 0.68$ & $\textbf{17.00} \pm 0.79$ \\
 MIX                 & $12.47 \pm 0.69$ & $17.66 \pm 0.80$ \\
 ORAL                & $12.43 \pm 0.69$ & $17.84 \pm 0.80$ \\
 ORAL\_NB            & $12.24 \pm 0.68$ & $17.67 \pm 0.80$ \\ \bottomrule
\end{tabular}
\label{tab:slu-results}
\end{table}

We can see that all the FlauBERT-O uncased models (FT, MIX, ORAL and ORAL\_NB) achieve comparable   results with the baseline FlauBERT models (cased and uncased).
It can be noticed that the FT model obtains an absolute improvement of 0.42\% CER compared to FlauBERT uncased. 
This suggests that pretraining LMs on ASR transcripts can benefit downstream tasks even though the domain is quite different (TV shows vs telephone conversations).
Although we use ASR transcripts with lower quality compared to \cite{Ghannay2021} we obtain almost similar CER  using the FT model.


One limitation of our experiments is that, even though SLU is applied to (noisy) ASR transcripts, during fine-tuning SLU models are updated using the (error free) reference transcripts of MEDIA. We believe that fine-tuning them on (noisy) ASR output could lead to even better results as it would match pre-training and test conditions. This improvement is left for future work but the next section will propose a classification task where ASR outputs are used during both fine-tuning and evaluation.



\section{Automatic Classification of TV Shows}
This section describes the evaluation of the different LMs on news classification. As part of the collection process, INA's documentalists finely segment newscasts of the main generalist TV/radio channels, 
and annotate them with content descriptions. This very rich metadata is used in particular to establish quantitative studies on the news in France. The InaStat barometer\footnote{\url{http://www.inatheque.fr/publications-evenements/ina-stat/ina-stat-sommaire.html}} has set up a stable methodology over time to classify news items into 14 categories (such as society, French politics, sport or environment). For classification experiments, we use the news items of 4 channels (TF1, France 2, France 3 and M6) for the years 2017, 2018 and 2019, which gives a total of 47,867 short TV shows, running on average 92 seconds.

The objective is to assess classification accuracy into the 14 categories solely on the basis of ASR transcripts. We compare results to a simple SVM classifier (with a non-parametric triangular kernel) on TF-IDF vectors baseline, with two vocabulary sizes of 5K and 20K words. A FlauBERT classifier is created for each model by appending a classification layer after mean pooling.
Models are not selected and rather we report results for different number of training epochs. Since the categories are not well-balanced, we use weighted F1  to evaluate performance. The experiments are systematically performed on 10 different random splits of the dataset, taking into account the cardinality of the 14 categories, so as to have 38K examples for the training set and 5K for the test set. Mean performance results and standard deviation are shown in Table \ref{tab:classif_split_asr_01_38k}.


\begin{table}[th]
\centering
\caption{TV news classification task: train 38K, test 5K, F1 weighted metric. Higher is better. }
\tabcolsep=1.60mm 
\begin{tabular}{@{}lccc@{}}
    \toprule
    \bf Model      & \bf Epoch 1    
                   & \bf Epoch 3
                   & \bf Epoch 10 \\
    \midrule
    SVM-20K            & \multicolumn{3}{c}{~~$0.777\pm0.005$} \\
    \midrule
    FBU             & $0.780\pm0.012$ & $0.807\pm0.010$ & $0.809\pm0.006$ \\
    \midrule
    FT              & $0.784\pm0.006$ & $0.807\pm0.006$ & $0.817\pm0.004$ \\
    MIX             & $0.791\pm0.005$ & $0.812\pm0.003$ & $0.820\pm0.004$ \\
    ORAL            & $\textbf{0.811}\pm0.004$ & $\textbf{0.829}\pm0.005$ & $\textbf{0.824}\pm0.004$ \\
    ORAL\_NB        & $0.798\pm0.004$ & $0.814\pm0.005$ & $0.824\pm0.004$ \\
    \bottomrule
\end{tabular}
\label{tab:classif_split_asr_01_38k}
\end{table}

If we look at the performance at the first epoch,  the FBU (flaubert\_base\_uncased) model has almost equivalent performance to the SVM baseline (0.777). It is only after a few iterations of fine-tuning that the model fits the ASR data and reaches 0.809. On the other hand, the models that have already seen ASR data during  training have better performance from the first epoch. The model trained only on ASR data is the best (ORAL). The ORAL\_NB model is slightly worse: in this model, only the tokenizer is different, so it seems better to not  adapt BPE to ASR transcripts, probably because it suffers from the closed vocabulary of the underlying system.
After 10 epochs, all FlauBERT-Oral models converge to the same performance and are better than FBU for this task.

\section{Syntactic Parsing of Spoken Dialogs}


In this section, the downstream task consists in jointly predicting part of speech tags (POS) and building a labelled dependency tree from speech transcripts.
We use our different LMs to generate contextual word representations and use them for predicting syntactic parses of speech transcripts.
We contrast the obtained results to a baseline model trained using FastText non-contextual representations,\footnote{\url{https://fasttext.cc}} and a model learning its own representations without any pretraining.

\subsection{The ORFEO dataset}

Our study relies on the annotated subset of the speech corpus of the ORFEO project \cite{benzitoun2016projet, nasr:hal-02973242}, gathered with the goal of reflecting the contemporary usage of the French language.
The audio recordings include work meetings, family dinners, narrations, political meetings, interviews, and goal-oriented telephone conversations.
Their duration varies from 4mn to 1h. 
The corpus is annotated with part-of-speech (POS) tags, lemmas, labeled dependency trees and sentences boundaries.
Annotations include 20 possible POS tags and 12 syntactic functions.
We randomly split the corpus into train/dev/test sets of respective sizes 134,716/27,937/29,529 words; we sampled from each source so that the various genres of speech are equally represented in each split.

\subsection{Parsing Model}

The syntax-predicting model is a transition based parser using the arc-eager transition system, which has been extended for the joint prediction of POS tags and parsing transitions \cite{dary-nasr-2021-reading}.
A single classifier is trained to predict transitions from features extracted from the current state of the analysis (word embebdings, predicted tags, history of transitions), for stack elements and a sliding window of size [-3;2] centered on the current word.
The POS sequence and dependency trees are greedily built from the sequence of transitions predicted by the classifier.
A dropout of 50\% is applied to the input vector, which passes through two hidden layers of size 3200 and 1600, both with 40\% of dropout and ReLU activations. The two decision layers are linear with a softmax.
Training runs for 40 epochs, using a dynamical oracle \cite{goldberg2012dynamic}, with scoring against the development set to select the best checkpoint.

\subsection{Experiments}

The first set of experiments compares input representations from the FlauBERT variants (FBU, FT, MIX, ORAL) to  non-contextual baselines. As with other experiments, LMs are applied on uncased transcripts without punctuation, which is not fair to the FBU model as it expects punctuation. Therefore we add a contrastive system with periods added at the end of each utterance prior to applying the LM (called FBU repunc), but added tokens are not passed to the parser, they just inform other embeddings.
Note that except for random initialized embeddings, token representations are not fine-tuned during training.

As pre-processing, we deanonymize the transcripts by replacing masked proper name tokens with non-ambiguous names randomly chosen for each recording. In the fasttext setting, representations are computed for unknown words from their character n-gram factors. Token representations are computed at the whole recording level in chunks of 512 tokens without overlap. The parser is applied on the reference transcript and reference segmentation. We use mean pooling for words that are split in multiple tokens by BPE.

Parsing performance is evaluated with Labeled Attachment Score (LAS), the accuracy of predicting the governor of each word and its dependency label, Unlabeled Attachment Score (UAS), which ignores dependency labels, and Part-of-speech tagging accuracy (UPOS) (scoring script from CoNLL campaigns).

Results presented in Table~\ref{t:syntax-main} show that pre-training is valuable for syntactic parsing in that setting and that pretraining on ASR-generated text (\textbf{FT}, \textbf{MIX} and \textbf{ORAL}) leads to a substantial improvement in LAS over the text-only FlauBERT model (\textbf{FBU}) even though there is no domain overlap between the TV shows on which the LMs are trained and the data of the ORFEO corpus. As in the previous task, we see no benefit in retraining BPEs (\textbf{ORAL\_NB}). Performance in the \textbf{FBU repunc} settings is on par with the best model, showing that absence of punctuation is an important factor affecting performance when applying text-only models, at least for predicting syntax. 

\begin{table}[htb]
\centering
\caption{Syntax prediction task: metrics are Labeled Attachment Score (LAS), Unlabeled Attachment Score (UAS) and Part-of-speech tagging accuracy (UPOS). Higher is better.}
\begin{tabular}{@{}lccc@{}}
\toprule
\bf Model &	\bf LAS &	\bf UAS &	\bf UPOS \\
\midrule
No pretrain & 	$84.92 \pm 0.44$ &	$88.48 \pm 0.37$ &	$94.51 \pm 0.29$ \\
Fasttext &	$85.36 \pm 0.10$ &	$88.76 \pm 0.05$ &	$95.12 \pm 0.03$ \\
\midrule
FBU &	$85.55 \pm 0.05$ &	$89.02 \pm 0.16$ &	$93.36 \pm 0.06$ \\
FBU repunc	& $87.48 \pm 0.28$	& $90.69 \pm 0.19$	& $95.03 \pm 0.03$ \\
\midrule
FT &	$86.81 \pm 0.09$ &	$90.22 \pm 0.06$ &	$94.99 \pm 0.10$ \\
MIX	& $86.33	\pm 0.15$ & $89.79	\pm 0.05$ & $94.43 \pm 0.29$ \\
ORAL	&  $\textbf{87.65}	\pm 0.11$ &  $\textbf{90.92}	\pm 0.10$ & $95.55 \pm 0.04$ \\
ORAL\_NB & $87.54 \pm 0.11$	& $90.73 \pm 0.08$ &  $\textbf{95.63} \pm 0.05$ \\
\bottomrule
\end{tabular}
\label{t:syntax-main}
\end{table}

\begin{table}[htb]
\centering
\caption{Differences in morphosyntactic performances, when parsing models of Table~\ref{t:syntax-main} are evaluated only on OOV words (OOV according to the automatic transcription system used to generate the text our LMs were trained on).}
\begin{tabular}{lrrr} \toprule									
{\bf Model} & {\bf $\Delta$LAS} & {\bf $\Delta$UAS} & {\bf $\Delta$UPOS} \\ \midrule
FBU	& -11.45 & -6.82 & -14.36 \\
MIX	& -11.93 & -7.33 & -14.07 \\
ORAL & {\bf -13.97} & {\bf -8.11} & {\bf -16.55} \\ \bottomrule
\end{tabular}
\label{t:syntax-oov}
\end{table}

Finally, in Table~\ref{t:syntax-oov} we compare the drop in parsing performance between three models, when they are evaluated only against OOV words (according to the ASR system).
Revealing that the \textbf{ORAL} model variants are more affected than the off-the-shelf FlauBERT baseline (\textbf{FBU}), despite their overall better performance (as seen in Table~\ref{t:syntax-main}).
This suggests future improvements of LMs trained on ASR-generated text if ASR output tokens were more consistent with LM tokens, and also suggests applying open vocabulary ASR.



\section{Conclusion and future work}

We investigated spoken language modeling using a massive amount of ASR-generated text (350,000 hours of diverse TV shows). Experiments suggest that, for the tasks we experimented with, systems relying on speech-informed LMs  have similar and better performance than systems relying on text-only LMs. 
The models resulting from this work for processing French speech are made available to the community. Deeper analysis of these improvements, on when and why models are better for spoken language  are items that are still to be addressed in the near future.

In this study, all  texts are uncased as our ASR only generates lowercased, unpunctuated transcripts. We believe that applying massively re-capitalisation and restoring punctuation might be beneficial to train stronger LMs for spoken language. We also believe that leveraging open vocabulary ASR and more diverse speech sources might yield more versatile LMs. A more fine-grain analysis on speech-only phenomena such as disfluencies is also needed.
Finally, some of the results obtained lead us to believe that it is important to further evaluate the impact of BPE units for spoken language modeling, as well as the consistency between ASR and LM tokens' vocabulary.

\section{Acknowledgements}
Some co-authors were funded by the AISSPER project supported by the French National Research Agency (ANR) under contract ANR-19-CE23-0004-01.


\FloatBarrier 
\bibliography{anthology,reduced}

\begin{thebibliography}{10}
\providecommand{\url}[1]{#1}
\csname url@samestyle\endcsname
\providecommand{\newblock}{\relax}
\providecommand{\bibinfo}[2]{#2}
\providecommand{\BIBentrySTDinterwordspacing}{\spaceskip=0pt\relax}
\providecommand{\BIBentryALTinterwordstretchfactor}{4}
\providecommand{\BIBentryALTinterwordspacing}{\spaceskip=\fontdimen2\font plus
\BIBentryALTinterwordstretchfactor\fontdimen3\font minus
  \fontdimen4\font\relax}
\providecommand{\BIBforeignlanguage}[2]{{%
\expandafter\ifx\csname l@#1\endcsname\relax
\typeout{** WARNING: IEEEtran.bst: No hyphenation pattern has been}%
\typeout{** loaded for the language `#1'. Using the pattern for}%
\typeout{** the default language instead.}%
\else
\language=\csname l@#1\endcsname
\fi
#2}}
\providecommand{\BIBdecl}{\relax}
\BIBdecl

\bibitem{koroteev2021bert}
M.~Koroteev, ``Bert: A review of applications in natural language processing
  and understanding,'' \emph{arXiv}, p. 2103, 2021.

\bibitem{qin2021survey}
L.~Qin, T.~Xie, W.~Che, and T.~Liu, ``A survey on spoken language
  understanding: Recent advances and new frontiers,'' in \emph{IJCAI}, 2021.

\bibitem{devlin2018bert}
J.~Devlin, M.-W. Chang, K.~Lee, and K.~Toutanova, ``Bert: Pre-training of deep
  bidirectional transformers for language understanding,'' \emph{arXiv preprint
  arXiv:1810.04805}, 2018.

\bibitem{shin2019effective}
J.~Shin, Y.~Lee, and K.~Jung, ``Effective sentence scoring method using bert
  for speech recognition,'' in \emph{Asian Conference on Machine
  Learning}.\hskip 1em plus 0.5em minus 0.4em\relax PMLR, 2019, pp. 1081--1093.

\bibitem{liu2020jointly}
C.~Liu, S.~Zhu, Z.~Zhao, R.~Cao, L.~Chen, and K.~Yu, ``Jointly encoding word
  confusion network and dialogue context with bert for spoken language
  understanding,'' \emph{Proc. Interspeech 2020}, pp. 871--875, 2020.

\bibitem{lai2020simple}
T.~M. Lai, Q.~H. Tran, T.~Bui, and D.~Kihara, ``A simple but effective bert
  model for dialog state tracking on resource-limited systems,'' in
  \emph{ICASSP 2020-2020 IEEE International Conference on Acoustics, Speech and
  Signal Processing (ICASSP)}.\hskip 1em plus 0.5em minus 0.4em\relax IEEE,
  2020, pp. 8034--8038.

\bibitem{ma2021extractive}
B.~Ma, H.~Sun, J.~Wang, Q.~Qi, and J.~Liao, ``Extractive dialogue summarization
  without annotation based on distantly supervised machine reading
  comprehension in customer service,'' \emph{IEEE/ACM Transactions on Audio,
  Speech, and Language Processing}, vol.~30, pp. 87--97, 2021.

\bibitem{siriwardhana2020jointly}
S.~Siriwardhana, A.~Reis, R.~Weerasekera, and S.~Nanayakkara, ``Jointly
  fine-tuning" bert-like" self supervised models to improve multimodal speech
  emotion recognition,'' \emph{arXiv preprint arXiv:2008.06682}, 2020.

\bibitem{ling2020bertphone}
S.~Ling, J.~Salazar, Y.~Liu, K.~Kirchhoff, and A.~Amazon, ``Bertphone:
  Phonetically-aware encoder representations for utterance-level speaker and
  language recognition,'' in \emph{Proc. Odyssey 2020 the speaker and language
  recognition workshop}, 2020, pp. 9--16.

\bibitem{baevski2020wav2vec}
A.~Baevski, Y.~Zhou, A.~Mohamed, and M.~Auli, ``wav2vec 2.0: A framework for
  self-supervised learning of speech representations,'' \emph{Advances in
  Neural Information Processing Systems}, vol.~33, pp. 12\,449--12\,460, 2020.

\bibitem{hsu2021hubert}
W.-N. Hsu, B.~Bolte, Y.-H.~H. Tsai, K.~Lakhotia, R.~Salakhutdinov, and
  A.~Mohamed, ``Hubert: Self-supervised speech representation learning by
  masked prediction of hidden units,'' \emph{IEEE/ACM Transactions on Audio,
  Speech, and Language Processing}, vol.~29, pp. 3451--3460, 2021.

\bibitem{kumar-etal-2021-bert}
A.~Kumar, M.~Narayanan~Sundararaman, and J.~Vepa, ``What {BERT} based language
  model learns in spoken transcripts: An empirical study,'' in
  \emph{BlackboxNLP Workshop on Analyzing and Interpreting Neural Networks for
  NLP}.\hskip 1em plus 0.5em minus 0.4em\relax ACL, Nov. 2021, pp. 322--336.

\bibitem{povey2011kaldi}
D.~Povey, A.~Ghoshal, G.~Boulianne, L.~Burget, O.~Glembek, N.~Goel,
  M.~Hannemann, P.~Motlicek, Y.~Qian, P.~Schwarz \emph{et~al.}, ``The kaldi
  speech recognition toolkit,'' IEEE Signal Processing Society, Tech. Rep.,
  2011.

\bibitem{galliano2009ester}
S.~Galliano, G.~Gravier, and L.~Chaubard, ``The ester 2 evaluation campaign for
  the rich transcription of french radio broadcasts,'' in \emph{Tenth Annual
  Conference of the International Speech Communication Association}, 2009.

\bibitem{giraudel2012repere}
A.~Giraudel, M.~Carr{\'e}, V.~Mapelli, J.~Kahn, O.~Galibert, and L.~Quintard,
  ``The repere corpus: a multimodal corpus for person recognition.'' in
  \emph{LREC}, 2012, pp. 1102--1107.

\bibitem{goryainova2014morpho}
M.~Goryainova, C.~Grouin, S.~Rosset, and I.~Vasilescu, ``Morpho-syntactic study
  of errors from speech recognition system.'' in \emph{LREC}, vol.~14, 2014,
  pp. 3050--3056.

\bibitem{deleglise2009improvements}
P.~Del{\'e}glise, Y.~Esteve, S.~Meignier, and T.~Merlin, ``Improvements to the
  lium french asr system based on cmu sphinx: what helps to significantly
  reduce the word error rate?'' in \emph{Tenth Annual Conference of the
  International Speech Communication Association}, 2009.

\bibitem{meignier2010lium}
S.~Meignier and T.~Merlin, ``Lium spkdiarization: an open source toolkit for
  diarization,'' in \emph{CMU SPUD Workshop}, 2010.

\bibitem{flaubert}
H.~Le, L.~Vial, J.~Frej, V.~Segonne, M.~Coavoux, B.~Lecouteux, A.~Allauzen,
  B.~Crabb{\'e}, L.~Besacier, and D.~Schwab,
  ``\BIBforeignlanguage{English}{{F}lau{BERT}: Unsupervised language model
  pre-training for {F}rench},'' in
  \emph{\BIBforeignlanguage{English}{LREC}}.\hskip 1em plus 0.5em minus
  0.4em\relax Marseille, France: ELRA, May 2020, pp. 2479--2490.

\bibitem{DBLP:journals/corr/abs-1907-11692}
Y.~Liu, M.~Ott, N.~Goyal, J.~Du, M.~Joshi, D.~Chen, O.~Levy, M.~Lewis,
  L.~Zettlemoyer, and V.~Stoyanov, ``Roberta: {A} robustly optimized {BERT}
  pretraining approach,'' \emph{CoRR}, vol. abs/1907.11692, 2019.

\bibitem{otmedia}
N.~Hervé, ``{{OTMedia}}, the {{TransMedia}} news observatory,'' in
  \emph{{{FIAT}}/{{IFTA}} Media Management Seminar 2019}, 05 2019.

\bibitem{TurDeMori2011}
G.~Tur and R.~D. Mori, ``Chapter 1: Spoken language understanding for
  human/machine interactions,'' in \emph{Spoken Language Understanding: Systems
  for Extracting Semantic Information from Speech}, 2011.

\bibitem{6424196}
F.~Morbini, K.~Audhkhasi, R.~Artstein, M.~Van~Segbroeck, K.~Sagae, P.~Georgiou,
  D.~R. Traum, and S.~Narayanan, ``A reranking approach for recognition and
  classification of speech input in conversational dialogue systems,'' in
  \emph{2012 IEEE Spoken Language Technology Workshop (SLT)}, 2012, pp. 49--54.

\bibitem{6998838}
G.~Mesnil, Y.~Dauphin, K.~Yao, Y.~Bengio, L.~Deng, D.~Hakkani-Tur, X.~He,
  L.~Heck, G.~Tur, D.~Yu, and G.~Zweig, ``Using recurrent neural networks for
  slot filling in spoken language understanding,'' \emph{IEEE/ACM Transactions
  on Audio, Speech, and Language Processing}, vol.~23, no.~3, pp. 530--539,
  2015.

\bibitem{Ghannay2021}
S.~Ghannay, A.~Caubri{\`{e}}re, S.~Mdhaffar, G.~Laperri{\`{e}}re, B.~Jabaian,
  and Y.~Est{\`{e}}ve, ``{Where Are We in Semantic Concept Extraction for
  Spoken Language Understanding?}'' \emph{Lecture Notes in Computer Science},
  vol. 12997 LNAI, pp. 202--213, 2021.

\bibitem{DBLP:conf/icassp/DinarelliKJB20}
M.~Dinarelli, N.~Kapoor, B.~Jabaian, and L.~Besacier, ``A data efficient
  end-to-end spoken language understanding architecture,'' in \emph{2020 {IEEE}
  International Conference on Acoustics, Speech and Signal Processing, {ICASSP}
  2020, Barcelona, Spain, May 4-8, 2020}.\hskip 1em plus 0.5em minus
  0.4em\relax {IEEE}, 2020, pp. 8519--8523.

\bibitem{Pelloin2021}
V.~Pelloin, N.~Camelin, A.~Laurent, R.~De~Mori, A.~Caubrière, Y.~Estève, and
  S.~Meignier, ``End2end acoustic to semantic transduction,'' in \emph{IEEE
  International Conference on Acoustics, Speech and Signal Processing
  (ICASSP)}, 2021.

\bibitem{Serdyuk2018}
D.~Serdyuk, Y.~Wang, C.~Fuegen, A.~Kumar, B.~Liu, and Y.~Bengio, ``{Towards
  End-to-end Spoken Language Understanding},'' \emph{ICASSP}, 2018.

\bibitem{BonneauMaynard2005}
H.~Bonneau-Maynard, S.~Rosset, C.~Ayache, A.~Kuhn, and D.~Mostefa, ``Semantic
  annotation of the french media dialog corpus,'' in \emph{INTERSPEECH}, 2005.

\bibitem{hahn2010comparing}
S.~Hahn, M.~Dinarelli, C.~Raymond, F.~Lefevre, P.~Lehnen, R.~De~Mori,
  A.~Moschitti, H.~Ney, and G.~Riccardi, ``Comparing stochastic approaches to
  spoken language understanding in multiple languages,'' \emph{IEEE
  Transactions on Audio, Speech, and Language Processing}, vol.~19, no.~6, pp.
  1569--1583, 2010.

\bibitem{NEURIPS2020_92d1e1eb}
A.~Baevski, Y.~Zhou, A.~Mohamed, and M.~Auli, ``wav2vec 2.0: A framework for
  self-supervised learning of speech representations,'' in \emph{Advances in
  Neural Information Processing Systems}, H.~Larochelle, M.~Ranzato,
  R.~Hadsell, M.~F. Balcan, and H.~Lin, Eds., vol.~33.\hskip 1em plus 0.5em
  minus 0.4em\relax Curran Associates, Inc., 2020, pp. 12\,449--12\,460.

\bibitem{ardila-etal-2020-common}
R.~Ardila, M.~Branson, K.~Davis, M.~Kohler, J.~Meyer, M.~Henretty, R.~Morais,
  L.~Saunders, F.~Tyers, and G.~Weber, ``\BIBforeignlanguage{English}{Common
  voice: A massively-multilingual speech corpus},'' in
  \emph{\BIBforeignlanguage{English}{LREC}}.\hskip 1em plus 0.5em minus
  0.4em\relax Marseille, France: ELRA, May 2020, pp. 4218--4222.

\bibitem{martin-etal-2020-camembert}
L.~Martin, B.~Muller, P.~J. Ortiz~Su{\'a}rez, Y.~Dupont, L.~Romary,
  {\'E}.~de~la Clergerie, D.~Seddah, and B.~Sagot, ``{C}amem{BERT}: a tasty
  {F}rench language model,'' in \emph{Proceedings of the 58th Annual Meeting of
  the Association for Computational Linguistics}.\hskip 1em plus 0.5em minus
  0.4em\relax Online: ACL, Jul. 2020, pp. 7203--7219.

\bibitem{benzitoun2016projet}
C.~Benzitoun, J.-M. Debaisieux, and H.-J. Deulofeu, ``Le projet orf{\'e}o: un
  corpus d’{\'e}tude pour le fran{\c{c}}ais contemporain,'' \emph{Corpus},
  no.~15, 2016.

\bibitem{nasr:hal-02973242}
A.~Nasr, F.~Dary, F.~Bechet, and B.~Favre, ``{Annotation syntaxique automatique
  de la partie orale du C{\'E}FC},'' \emph{{Langages}}, Sep. 2020.

\bibitem{dary-nasr-2021-reading}
F.~Dary and A.~Nasr, ``The reading machine: A versatile framework for studying
  incremental parsing strategies,'' in \emph{Proceedings of IWPT 2021}.\hskip
  1em plus 0.5em minus 0.4em\relax Online: ACL, Aug. 2021, pp. 26--37.

\bibitem{goldberg2012dynamic}
Y.~Goldberg and J.~Nivre, ``A dynamic oracle for arc-eager dependency
  parsing,'' in \emph{Proceedings of COLING 2012}, 2012, pp. 959--976.

\end{thebibliography}
\bibliographystyle{IEEEtran}

\end{document}